# CLOSING THE LOOP: FORMALLY VERIFIED LAW AS A REWARD SIGNAL FOR SELF-IMPROVING LEGAL AI

Armin Heydari[*] and Torben Leowald[†‡]

April 12, 2026

TABLE OF CONTENTS

*Our companion article argued that scaling transformer-based language models alone will not produce autonomous legal reasoners. The deeper difficulty, we suggested, is that contemporary training methods depend on verifiable outcomes, and legal reasoning has historically resisted the formal structure that makes verification possible. This article develops the architecture that supplies it, adapting the "LLM proposes, verifier disposes" paradigm from mathematical AI to the distinctive demands of law. We present an architecture comprising LLM-driven autoformalization into a formal legal calculus extending Catala, a verification kernel, and explanation generation grounded in formal proof traces. For the computational components of law, the architecture provides provable correctness. For open-textured legal analysis, it provides structural guarantees: every required stage of the legal argument is addressed, argumentation is exercised at the correct stages and not omitted, and the deductive links between steps are valid. We demonstrate the architecture on procedural deadline calculations in German law, Commerce Clause analysis in U.S. constitutional law, and cross-jurisdictional sanction proportionality. We further show that the same architecture has a structural advantage for legal AI training: a deterministic external verifier supplies verifiable outcomes for legal problems and thereby closes the traditional reinforcement-learning loop gap in law.*

---

[*] Department of Philosophy, Harvard University, Cambridge, MA, USA.

[†] Columbia Law School, Columbia University, New York, NY, USA.

[‡] The large language models Claude Opus 4.6 (Anthropic, 2026) and GPT-5.3 and GPT-5.4 (OpenAI, 2026) assisted in editing, research organization, literature search, and formatting.

## I — INTRODUCTION

In the companion article, we defended three claims: first, that current legal AI systems are useful augmentation tools but not autonomous legal reasoners; second, that the empirical failures of these systems track deeper architectural limits of transformer-based language models on compositional reasoning; and third, that law demands provable verification guarantees, not merely high accuracy. The question that the companion's argument left open was whether any architecture could satisfy law's demand for accountable correctness.

This article argues that it can. The "LLM proposes, verifier disposes" architecture, adapted from applications of generative artificial intelligence in mathematics, is the most plausible path toward a legal AI system that can reliably certify its outputs and, ultimately, legal superintelligence. The core idea is that generation can be probabilistic as long as verification is deterministic. The language model proposes candidate legal analyses; a formal verifier accepts or rejects each step against a rigorous standard. Nothing that fails verification reaches the final output. Verification thereby also supplies the reward signal needed for the reinforcement learning used to train the models in the first place.

By legal superintelligence we mean a system that can generate a legally correct theory and argument for any legal question insofar as the question has a legally determinate answer, and that can identify the points of genuine legal contestation when it does not. More concretely, it is a system that can reason over law at a depth and breadth exceeding the capacity of any individual human lawyer or reasonably sized team, while remaining at least as auditable and trustworthy as a human lawyer at every step of its reasoning. When we say *auditable*, we mean that the system exposes a reasoning trace that allows a human reviewer to inspect, at each step, which legal rule was applied, to which facts, with which result. This implies that the reviewer can identify precisely where, if anywhere, the system exercised a judgment that the reviewer may wish to contest. By *trustworthy*, we mean that the system's outputs satisfy the following condition: for the parts of the analysis amenable to formal verification (e.g., deadline calculation), the correctness of the result is guaranteed; for the parts requiring judgment, the points at which judgment was exercised are explicitly identified. Our definition therefore neither requires that legal superintelligence, if it exists, is always right (just as two human legal experts can reasonably disagree over the conclusion of a case) nor that it replaces lawyers (since the human and cultural frictions that create the need for law in the first place are not replaced by perfect reasoning).

These notions at hand, we can reframe the results from the companion paper as the thesis that current legal AI systems cannot achieve legal superintelligence because their reasoning is neither auditable nor trustworthy. It is not auditable because the models' reasoning paths are opaque, even in the presence of chain-of-thought prompting. And they are not trustworthy because they have no satisfactory mechanism for distinguishing formally verified steps from perhaps highly accurate probabilistic guesses, even in the presence of retrieval-augmented generation.

The architecture that we propose is inspired by two developments outside law. First, in mathematics, the "LLM proposes, verifier disposes" paradigm has achieved competition-level performance. Google DeepMind's AlphaProof, combined with AlphaGeometry 2, achieved silver-medal-level performance at the 2024 International Mathematical Olympiad by coupling a Gemini-based language model with reinforcement learning within the Lean 4 proof assistant.[1] DeepSeek-Prover-V2 achieved an 88.9% pass ratio on the MiniF2F-test benchmark.[2] Harmonic's Aristotle platform solved five of six problems at the 2025 IMO with fully verified Lean 4 proofs, which is a gold-medal-equivalent performance.[3] In each case, the basic architecture is this: a language model generates candidate proof steps rapidly and with creative breadth; a proof assistant accepts or rejects each step with deductive certainty; and reinforcement learning uses the verifier's signal to improve the proposer.

It is possible to draw a parallel to law. Mathematical statements correspond to legal claims; formal logics correspond to legal domain-specific languages; proof checking corresponds to argument verification. Of course, there is one big disanalogy. Most of the mathematical statements that mathematical practice seeks to prove or disprove are either verifiably provable or verifiably disprovable. Legal questions, by contrast, often admit of genuinely competing interpretations. This is the problem of *open texture*, which we address throughout the article.

We proceed in five parts. Part II presents the proposed architecture. Part III demonstrates the architecture on three legal problems of increasing complexity. We address objections in Part IV before concluding in Part V.

## II — ARCHITECTURE

The basic insight is that generation can be probabilistic as long as verification is deterministic. In the mathematical AI systems described above, the language model – a stochastic system optimized for plausible text generation – generates candidate proof steps. The proof assistant – a deterministic system formalizing mathematics – accepts or rejects each step. The verifier's acceptance guarantees that a proposed proof is correct, while the verifier's rejection tells the language model to try again. The legal analog is as follows. A language model translates natural-language statutes, regulations, and precedent into a typed formal language, a process known as *autoformalization*.[4] The resulting formal program is submitted to a verification kernel that checks it for well-typedness (that is, that every operation in the program is applied to the right kind of object: avoiding category errors such as citing a remedies case for a jurisdictional proposition) and the satisfaction of certain asserted constraints. If the program passes verification, its outputs are guaranteed to follow from the formalized rules. If it fails, the error is reported to the language model as structured feedback for a second attempt.

For that reason, the verifier is not only a deployment-time gatekeeper; it is also a source of the training reward.[5] First construct a formal artifact that an independent system can check; then use the result of that check to improve the proposer. Formalization and verification are therefore part of one model training pipeline, allowing for recursive improvement in legal AI models.

[1] Thomas Hubert et al., Olympiad-Level Formal Mathematical Reasoning with Reinforcement Learning, 651 Nature 607 (2026).

[2] Ren et al., *DeepSeek-Prover-V2: Advancing Formal Mathematical Reasoning via Reinforcement Learning for Subgoal Decomposition*, arXiv:2504.21801 (2025).

[3] Tudor Achim et al., Aristotle: IMO-Level Automated Theorem Proving, arXiv:2510.01346 (2025).

[4] On autoformalization generally, see Yuhuai Wu et al., Autoformalization with Large Language Models, NeurIPS (2022). On autoformalization as a grounding mechanism for quantitative reasoning, see Jin Peng Zhou et al., Don't Trust: Verify—Grounding LLM Quantitative Reasoning with Autoformalization, ICLR (2024). On autoformalization in the legal domain specifically, see Gabriele Lorenzo, Aldo Pietromatera & Nils Holzenberger, Translating Tax Law to Code with LLMs: A Benchmark and Evaluation Framework, in Proceedings of the Natural Legal Language Processing Workshop 2025, at 31 (2025).

[5] See Hubert et al., supra note 1; Ren et al., supra note 2.

A. The Problem of Open Texture

We must now confront the problem of open texture. Whereas most of the mathematical statements that mathematical practice seeks to prove or disprove are both formally representable and either verifiably provable or verifiably disprovable, legal questions often admit of genuinely competing interpretations. This means that legal arguments may be neither verifiably provable nor disprovable, and there may not even be a uniquely correct formalization of the relevant legal arguments.

We argue that this worry does not harm the promise of formally verified legal AI if either one of two assumptions is warranted. The first is that a meaningful portion of the law concerns *bound determinations*, where there is no room for competing interpretations. Examples may include deadline calculations, tax computations, eligibility rules, and statutory cross-references. If this assumption is true, formally verified legal AI is promising for the practice and development of law at least insofar as bound determinations are important therein. Alternatively, one may assume that while there is a meaningful portion of the law that concerns open-textured analysis (e.g., constitutional balancing tests, proportionality assessments, and reasonableness standards), such open-textured analysis carries formally verifiable desiderata of quality (e.g., that a constitutional balancing test must discuss all of the affected rights). Under this assumption, formally verified legal AI is promising for the practice and development of law at least insofar as it enables the *structural verification* of open-textured analysis: e.g., that the argument addresses all required factors, in the correct order, with valid deductive links between steps.

We find both assumptions plausible. They also map to well-known distinctions in the literature. H.L.A. Hart argued that legal rules expressed in general language have a "core of settled meaning" where application is clear and a "penumbra of doubt" where it is not.[6] Hart's canonical example – "no vehicles in the park" – clearly covers automobiles but becomes uncertain for toy cars, bicycles, or war-memorial tanks. The penumbra, Hart argues, requires judges to exercise discretion. Robert Alexy draws a similar distinction between *subsumption* and *balancing*.[7] The subsumption formula is a deductive chain: a major premise states a rule, a minor premise classifies the facts, and a conclusion follows by modus ponens. The weight formula for principle-balancing, by contrast, requires gradated judgments about the intensity of interference with competing principles. These are judgments that do not follow deductively from any antecedent premises, but rather judgments requiring judges to exercise discretion. Finally, Neil MacCormick distinguishes *first-order justification* – the legal syllogism applied in clear cases – from *second-order justification* required when first-order reasoning fails.[8] Bound determinations, then, correspond to Hart's core, Alexy's subsumption, and MacCormick's first-order justification, while open-textured analysis corresponds to Hart's penumbra, Alexy's balancing, and MacCormick's second-order justification.

[6] H.L.A. Hart, The Concept of Law 119–32 (1961).

[7] Robert Alexy, *On Balancing and Subsumption: A Structural Comparison*, 16 Ratio Juris 433 (2003).

[8] Neil MacCormick, *Legal Reasoning and Legal Theory* (1978).

Of course, the picture becomes muddier as one considers less positivist views about the law. On realist views such as Oliver Wendell Holmes's, "the felt necessities of the time, the prevalent moral and political theories, intuitions of public policy, avowed or unconscious, even the prejudices which judges share with their fellow-men, have had a good deal more to do than the syllogism in determining the rules by which men should be governed."[9] This appears to imply that formally verifying legal arguments is a futile endeavor, since it attempts to formalize law by formalizing precedent, tests, statutes, commentary, and the like, despite legal rules actually being determined by the moral and political theories of judges, which are not being formalized. On interpretivist views, such as Ronald Dworkin's, the law is not determined by abstract rules but by *interpretation*, where interpretation concerns normative objects that justify the institutional practice of law.[10] Again, formalizing legal rules as abstract objects would appear to miss the mark by disregarding normative objects and their role in justifying the institutional practice of law.

This appearance, however, is deceptive. For one, the mere fact that legal rules are determined by moral and political theories of judges does not mean that there is no suitable formalization of the thus-determined legal rules. Much legal practice is determinate once the facts of the case and their similarity to patterns of facts in the case law are settled. Second, although formal objects corresponding to rules are indeed no normative objects, *in designing* a suitable formalization of the law, the law can be determined through interpretation and, thus, through consideration of normative objects that justify the institutional practice of law. For example, commitments to individual rights justify an institutional practice of assessing the proportionality of rights infringements with respect to all affected persons; this commitment, in turn, can inform the choice that a suitable formalization of the law should take individuals, as opposed to, say, collectives, as primitive entities.

Our proposed architecture therefore deals with the problem of open texture by delivering two tiers of guarantee. For bound determinations, the architecture provides provable correctness for the final result. The system certifies, with deductive certainty, that the computation follows from the formalized rules. For open-textured analysis, the system certifies that the argument addresses every required step with valid deductive links between steps. Reasonable people, and the system itself, may still disagree about the premises.

Over time, these two levels may converge. John Horty, developing a proposal of Grant Lamond, argues that the appeal to a settled semantic core is dispensable: what determines whether a court is required, permitted, or forbidden to apply a predicate in a given situation is not core meaning but a *priority ordering over reasons* that the relevant body of precedent has established.[11] Take Hart's example of the predicate "vehicle," in a bylaw barring vehicles from a park. Whether the term applies to a motorized scooter is, in Hart's terminology, a penumbral question. But suppose a court has already held that a moped is a prohibited vehicle, resting that holding on the reason that the moped is motor-powered and capable of speeds dangerous to pedestrians, in preference to the competing reason that it is used recreationally. That decision ranks the danger-and-speed reason above the recreational reason. When the motorized scooter presents that same danger-and-speed reason, the court's decision is bound by the priority of the danger-and-speed reason over the recreational reason that the previous decision laid out. The court is then required to classify the scooter as a vehicle. Had the new situation presented only the recreational, low-speed reason (e.g., a child's tricycle), the same body of precedent would forbid the classification, there being no reason present on which the prohibition could rest. And had it presented a reason the precedents had never weighed (e.g., an ambulance, whose emergency function bears on neither prior reason) the ordering would be silent, leaving the court permitted to decide either way, and in so deciding to extend the priority ordering over reasons for the cases that follow. The predicate's application is thus fixed, case by case, by the ordering precedent induces rather than by a semantic core; and as that ordering grows more complete, the band of situations in which application is merely permitted (the residue of open texture) narrows, while the band in which it is required or forbidden, and so amenable to structural verification, widens. Horty's account thereby shows that the parts of the law most amenable to structural verification and the parts of the law least amenable may at some point converge, because the residue of open texture narrows with the proliferation of precedent.[12]

[9] Oliver Wendell Holmes, Jr., The Common Law 1 (1881).

[10] Ronald Dworkin, *Law's Empire* (1986).

[11] See John Horty, *Rules and Reasons in the Theory of Precedent*, 17 Legal Theory 1 (2011); John Horty, The Logic of Precedent: Constraint, Freedom, and Common Law Reasoning (2025). Horty develops a proposal of Grant Lamond, *Do Precedents Create Rules?*, 11 Legal Theory 1 (2005).

B. Layer 1: Autoformalization

The first layer of the architecture translates natural-language legal provisions into a typed formal language we call $\lambda_{law}$, which extends the Catala programming language with deontic modalities and temporal operators.

Catala is a programming language for the law, introduced by Merigoux, Chataing, and Protzenko.[13] Its theoretical foundation is the observation that statutes characteristically state general rules and then add exceptions. Catala captures this as a *default expression* $\langle e_1, \ldots, e_n \mid j :- c \rangle$ with the intended meaning that if justification $j$ is true, the term should evaluate to $c$ unless one of the exceptions $e_1, \ldots, e_n$ applies. More precisely: if exactly one exception yields a value, return that value; if two or more yield values, return a *conflict error* $\bot$; if none yields a value and the justification is true, return the consequent; if the justification is false, return an *empty error* $\emptyset$. The intended legal interpretation is that a value is a determinate conclusion; a conflict signals that the statute contains an ambiguity requiring open-textured resolution; and emptiness signals a gap in the statutory scheme. Catala has already helped discover a real bug in the official implementation used by France's Caisse d'Allocations Familiales: the production software for family benefits calculations diverged from legislative intent on fine-grained distinctions that only became visible through formalization.

[12] Note that we take Horty's case, which is likely the most comprehensive formal model of the law in the literature, to strengthen our case that formalizing law is a realistic endeavor. Apart from this conclusion, our formal model diverges significantly from Horty's: while Horty takes *reasons* as primitives and formalizes law in terms of priority orderings over such reasons, we take *rules* as primitives, expecting that these will be better suited to training language models of the law.

[13] Denis Merigoux, Nicolas Chataing & Jonathan Protzenko, Catala: A Programming Language for the Law, 5 Proc. ACM Programming Languages, no. ICFP, art. 77, at 1 (2021).

$\lambda_{law}$ extends Catala's default calculus in two directions. First, it adds deontic type constructors O(τ), P(τ), F(τ) for obligation, permission, and prohibition; these will be useful for cross-jurisdictional legal questions (e.g., comparing two sanctions regimes, one of which prohibits and the other permits a transaction). Derived obligations serve as inputs for further normative computation, modeled by a *reusable output* operation adapted from input/output logics.[14] This allows the system, for example, to model that from the fact that condition *a* triggers obligation *x* and that *x* together with *a* triggers obligation *y*, it follows that *y* is also obligatory. Second, $\lambda_{law}$ adds temporal operators for date arithmetic with explicit rounding semantics, following Monat, Fromherz, and Merigoux's formalization of date arithmetic for the law.[15]

In practice, a large language model can translate natural-language statutes, regulations, precedent, and legal rules into $\lambda_{law}$ programs. The feasibility of this translation has been empirically established. Lorenzo, Pietromatera, and Holzenberger achieved 88.8% syntactically valid Catala code using GPT-4.1 with few-shot prompting on French tax law.[16] Outside of legal applications, researchers have already achieved 99% soundness by using redundant LLM formalizations cross-checked for semantic equivalence via symbolic reasoning.[17]

Crucially, the correctness of the autoformalization is not *assumed* in our proposed architecture. It is verified by Layer 2. The autoformalization is a stochastic map from natural-language legal text to $\lambda_{law}$ programs, and the system treats it as such.

C. Layer 2: The Verification Kernel

The *kernel* is a small, trusted computing base that is independent of the language models employed in Layer 1. It type-checks $\lambda_{law}$ programs (ensuring no category mistakes are encoded), performs static analyses (for example, that every input produces a defined output or that the program is conflict-free and without temporal ambiguities), and verifies that certain asserted constraints are not violated. The latter is done by negating the relevant constraint and submitting it to an *SMT solver* such as Z3.[18] An SMT solver (Satisfiability Modulo Theories solver) determines whether a given logical formula is satisfiable, that is, whether there exists an assignment of values to its variables that makes the formula true. If the negation of the constraint is unsatisfiable – if there is no input that violates it – then the constraint holds for all inputs. For example, the invariant "The amount of tax that is owed is greater than or equal to zero." is verified by asking the solver whether there exists any input for which the computed tax is negative. If the solver reports that no such input exists, the invariant is proven. If it finds a violating input, it returns a counterexample.[19]

---

[14] David Makinson & Leendert van der Torre, Input/Output Logics, 29 J. Phil. Logic 383 (2000).

[15] Raphaël Monat, Aymeric Fromherz & Denis Merigoux, Formalizing Date Arithmetic and Statically Detecting Ambiguities for the Law, ESOP (2) 2024, at 421.

[16] Lorenzo, Pietromatera & Holzenberger, *supra* note 4.

[17] Samuel Bayless et al., A Neurosymbolic Approach to NL Formalization and Verification, arXiv:2511.09008 (2025).

[18] Leonardo de Moura & Nikolaj Bjørner, Z3: An Efficient SMT Solver, TACAS 2008, at 337.

[19] The kernel's reliance on an SMT solver to discharge such constraints has an analog in the formal theory of precedent. Horty notes that verifying that a body of precedent is consistent (i.e., that no pair of opposing reasons is each ranked above the other) requires a finite search over combinations of reason-pairs, for m factors favoring one side and n the other (*The Logic of Precedent*, *supra* note 11, §2.2.1). Making such verifications is exactly the task SMT solvers are built to automate.

The kernel is trustworthy for three reasons. First, it is small: a small codebase minimizes the attack surface for bugs, in contrast to a language model with hundreds of billions of parameters. Second, it is LLM-independent: the language model cannot influence the checking, since the kernel operates on the formal program alone. Third, it is backed by the soundness of the SMT solver's procedures: Z3's decision procedures for quantifier-free linear integer arithmetic and quantifier-free linear real arithmetic are sound: if the solver reports unsatisfiability, the formula is genuinely unsatisfiable. This is analogous to the Lean type checker in mathematical AI: Lean's kernel is a small, trusted program whose correctness can be independently verified, and its acceptance of a proof constitutes a guarantee of correctness.

D. Layer 3: Explanation Generation

Every conclusion of the kernel is accompanied by a proof trace: a sequence of triples ⟨natural-language explanation, statutory provision or precedent, verified intermediate value⟩. The LLM then generates a natural-language explanation of the proof trace. Since this happens after the kernel has concluded its work, errors in explanation cannot affect the verified result.

For open-textured analysis, the triple can be adapted so that the rule component cites a precedent and its ratio decidendi, and the value component records the analogical mapping between the precedent's fact pattern and the case at hand. Verification then provides a weaker but still valuable guarantee: that the cited case exists, the quoted ratio is faithful, and the structural analogy is well-formed. Extending the kernel to support defeasible analogical reasoning in this way is an important direction for future work.

E. Training: Reinforcement Learning from Verifier Feedback (RLVF)

Reinforcement learning improves a model by rewarding outputs that satisfy an objective and penalizing those that do not. The difficulty in many domains is the reward itself: where correctness must be judged by a second model or by human raters, the signal is noisy, costly, and gameable, since a capable generator learns to satisfy the judge rather than the underlying standard. Our proposed architecture solves this problem through the kernel, which provides a perfect reward signal for training. We adapt GRPO (Group Relative Policy Optimization) – the algorithm behind DeepSeek-R1's reasoning capabilities[20] – for legal autoformalization. GRPO eliminates the critic network required by standard reinforcement learning algorithms such as PPO.[21] For each prompt, GRPO samples a group of candidate outputs from the current policy, computes rewards for each candidate, and estimates the advantage of each candidate as its z-score within the group. This group-based comparison replaces the learned value function, reducing memory requirements by approximately half and simplifying the training pipeline. For legal autoformalization, we propose a three-tier reward signal derived from the kernel: full marks if the $\lambda_{law}$ program type-checks and passes the test suite; partial credit if it type-checks but fails assertion verification; and zero if it does not type-check.

[20] DeepSeek-AI et al., *DeepSeek-R1: Incentivizing Reasoning Capability in LLMs via Reinforcement Learning*, arXiv:2501.12948 (2025).

[21] Shao et al., *DeepSeekMath: Pushing the Limits of Mathematical Reasoning in Open Language Models*, arXiv:2402.03300 (2024).

Test suites derive from legal commentary and worked examples. A consistency reward measures whether the code's scope-and-exception structure mirrors the statute's sections. This parallels DeepSeek-Prover-V2's use of Lean 4's binary accept/reject signal as the sole training reward.[22] As the human-formalized corpus grows, it becomes training data for increasingly accurate autoformalization.

Three features of legal autoformalization recommend GRPO in particular. The formalization targets are heterogeneous in difficulty, since encoding a filing-deadline rule and encoding a multi-jurisdictional sanctions interaction are not comparable tasks; an algorithm scoring each candidate against an absolute scale would be dominated by the easy cases, whereas GRPO's per-group normalization removes that distortion. The kernel emits a graded rather than a binary signal, which GRPO consumes directly. Finally, a learned value function is unreliable when estimated from the small expert corpus that legal formalization will occupy for some time, so its absence in GRPO is an advantage rather than a sacrifice. The three-tier signal can itself be refined into a graded reward mirroring the kernel's sequence of checks, weighting each verification stage in proportion to what it certifies.[23] The kernel thereby exposes a dense reward, each intermediate verdict independently meaningful, where most verifiable-reward systems expose only a terminal binary; this density is precisely what process supervision has had to purchase through expensive human step-labeling.[24]

One might ask why the expert corpus is not simply used as supervised training data, without verifier and reinforcement learning at all. Supervised fine-tuning is useful as a warm start, but it is insufficient as the main engine. A small corpus teaches the model to reproduce the examples it has seen; it does not reliably teach the compositional generalization needed to handle unfamiliar, nested, and exception-laden rules.[25] Verifier feedback supplies the missing leverage: the model can explore new formalizations, the kernel can score them, and GRPO can shift probability mass toward the candidates that survive.

The concrete pipeline we propose runs as follows. The process begins with a seed corpus of expert formalizations, each kernel-verified and paired with its natural-language source. For each source the policy samples a group of candidate λlaw programs; uniformly failing groups are resampled until they contain variance, and uniformly passing ones discarded. Each candidate is run through the kernel, the test suite, and the SMT assertions, checked for structural fidelity to the statute, and scored for faithfulness by back-translation against the source, with the reward assembled so that the binary kernel verdict fixes its sign while the graded components modulate only its magnitude. The policy is then updated by GRPO on the group-relative advantage, under a KL penalty and with the dynamic-sampling and clipping correctives that formal-domain reinforcement learning has converged upon. Failures at the model's capability frontier—groups of mixed outcome or high uncertainty—are routed to legal experts, whose corrections enlarge the corpus, the test suite, and the reference set; on convergence these are folded back in and the loop re-initialized from the improved policy. As the corpus of verified formal law grows, autoformalization grows more accurate, which enlarges the corpus in turn—the legal counterpart of the flywheel by which AlphaFold's accuracy on structures from the Protein Data Bank enabled generalization to novel proteins[26].

---

[22] Ren et al., *DeepSeek-Prover-V2: Advancing Formal Mathematical Reasoning via Reinforcement Learning for Subgoal Decomposition*, arXiv:2504.21801 (2025).

[23] By way of illustration only, the stages might be weighted well-typedness (0.1), static analysis (0.2), the assertion suite (0.4), SMT-discharged invariants (0.2), and faithful reproduction of the statute's scope-and-exception structure (0.1). The weights are tuned empirically; nothing in the argument turns on the particular figures.

[24] Lightman et al., *Let's Verify Step by Step*, ICLR (2024) (demonstrating the value of step-level process supervision while noting the cost of the human step labels it requires).

[25] On the limits of purely supervised learning for systematic compositional generalization, see Brenden M. Lake & Marco Baroni, Generalization Without Systematicity: On the Compositional Skills of Sequence-to-Sequence Recurrent Networks, ICML 2018; Daniel Keysers et al., Measuring Compositional Generalization: A Comprehensive Method on Realistic Data, ICLR 2020. On using reinforcement learning rather than SFT alone to elicit reasoning behavior, see DeepSeek-AI et al., supra note 20.

F. Limitations

First, the architecture does not guarantee the correctness of open-textured, premise-level reasoning. When the Commerce Clause analysis in Part III asks whether a particular activity is "economic in nature," the answer depends on a judgment that no formal system can supply. What the architecture guarantees is that the question was asked, that it was asked at the right stage of the analysis, and that the subsequent reasoning respects the answer: not that the answer itself is correct.

Second, the architecture does not guarantee that the formalization is correct. The $\lambda_{law}$ program might encode the wrong rule. This is the *specification gap*: the gap between the law and its formalization. We return to this in Part V. The gap is real but strictly smaller than for probabilistic systems, which face the same specification problem and also face the problems of hallucination, miscalculation, and opaque reasoning.

Third, we have not yet benchmarked the proposed architecture against non-verified legal AI architectures. That limitation should be explicit: this article provides the theoretical foundation for implementing a verifier-based legal-AI pipeline, not evidence that an implemented system already outperforms existing systems. Our theory rests on three grounds. First, mathematical precedent shows that verifier-in-the-loop reinforcement learning can carry formal reasoning systems to competition-level performance on a proposer-plus-checker architecture closely analogous to ours. Second, early legal evidence points in the same direction: Kant et al. show that guided neuro-symbolic formalization can outperform vanilla prompting on contract-coverage tasks, while unguided formalization can underperform even the unaided baseline; Lorenzo, Pietromatera, and Holzenberger likewise show that legal text can be translated into Catala code reliably enough to make legal-code generation and legal-code evaluation concrete benchmarkable tasks.[27] Third, the structural argument remains: reinforcement learning against a sound, deterministic verifier cannot reward an output that the verifier cannot certify. The reward is therefore non-gameable in the relevant sense. The remaining empirical question is whether the generator can be driven toward the region the verifier accepts; the mathematical results suggest that, for analogous verifier-backed domains, it can. The training data can be drawn from three sources of decreasing cost: an expert-formalized corpus, which is the scarcest input and the legal counterpart of the Protein Data Bank; rejection sampling once the model produces plausible λlaw; and expert iteration during reinforcement learning.

[26] Jumper et al., Highly Accurate Protein Structure Prediction with AlphaFold, 596 Nature 583 (2021). As AlphaFold's accuracy on structures drawn from the Protein Data Bank enabled prediction of novel structures, an enlarging corpus of kernel-verified λlaw is expected to improve autoformalization of rules the corpus does not yet contain.

[27] Manuj Kant, Sareh Nabi, Manav Kant, Roland Scharrer, Megan Ma & Marzieh Nabi, Towards Robust Legal Reasoning: Harnessing Logical LLMs in Law, arXiv:2502.17638 (2025) (reporting that guided neuro-symbolic analysis outperformed vanilla prompting on contract-coverage tasks while unguided LLM-generated logic often underperformed the vanilla baseline); Lorenzo, Pietromatera & Holzenberger, supra note 4.

### G. Differences to Extant Legal AI

We compare the proposed architecture to the dominant paradigm in legal AI: retrieval-augmented generation (RAG) with chain-of-thought prompting. RAG systems retrieve source documents and inject them into the language model's context window before generation. This reduces the frequency of fabricated citations, but it cannot guarantee that retrieved authority is controlling, timely, or supportive of the proposition asserted in the output.

Chain-of-thought prompting faces a structurally analogous limitation. Modern reasoning models produce markedly better intermediate steps than their predecessors. But no mechanism internal to the model verifies that a given step follows from the preceding one by a valid rule of inference rather than by a plausible textual continuation.

## III — EXAMPLES

### A. Procedural Deadlines

Missed deadlines account for 28% of malpractice claims and 34% of damages[28]. Yet the calculation is purely algorithmic.

As an example, German deadline calculation (*Fristberechnung*, §§186–193 BGB) has four components. First, the start of period (§187 BGB): an event-triggered period (*Ereignisfrist*) begins the day after the event; a date-triggered period (*Terminfrist*) begins on the named date. Second, the end of period (§188(2) BGB): the period ends on the day corresponding to the day preceding the trigger. Third, month-end rounding (§188(3) BGB): if the corresponding day does not exist (e.g., one month from January 31 yields no February 31), the period ends on the last day of that month. Fourth, extension (§193 BGB): deadlines falling on Sundays, public holidays, or Saturdays extend to the next working day.

Formalised into $\lambda_{law}$, the start-date computation directly instantiates the default operator. The base case is §187(1): skip the event day. The exception is §187(2): count the start day. Exactly one applies; the kernel verifies this statically. The raw deadline adds the statutory duration using $\lambda_{law}$'s temporal operators with rounding mode ↓ (matching §188(3)). The final deadline applies the working-day shift parameterized by jurisdiction.

The kernel verifies for all inputs that "deadline > trigger" and "deadline ≥ raw_deadline" and that the calculations are correct.

[28] Professional Liability Fund, inBrief: Malpractice Prevention Education for Oregon Lawyers, Issue 139 (Dec. 2019).

B. Commerce Clause Analysis

Congress may regulate under the Commerce Clause if the activity (1) uses channels of interstate commerce; (2) concerns instrumentalities, persons, or things of interstate commerce; or (3) constitutes an activity with a substantial effect on interstate commerce, subject to the *Lopez* five-factor test: (F1) Is the regulated activity economic in nature? (F2) Is there a jurisdictional hook tying the regulated conduct to interstate movement? (F3) Are there express legislative findings regarding the effect on interstate commerce? (F4) Is the causal link between activity and interstate commerce too attenuated? (F5) Does the regulation concern domains traditionally reserved to the states?[29]

Formalised into $\lambda_{\text{law}}$, the three-category test is a default expression: the base case is the substantial-effects branch; exceptions are the channels and instrumentalities categories, which, when applicable, resolve the question without reaching the five-factor test. The substantial-effects branch is itself a scope with five context variables, one per *Lopez* factor. Each factor is typed as enum{supports, opposes, neutral} with a mandatory natural-language justification field.

The kernel enforces three structural properties: that all five factors are addressed (no $\emptyset$ in the scope outputs); that they are addressed in the correct sequence (e.g., if the activity is not economic, the remaining factors cannot rescue it); and that they are evaluated consistently: if the jurisdictional-hook factor (F2) is "supports," the analysis should have terminated at Category 2, not reached the substantial-effects test. The substantive assessments of whether, for example, a particular activity is "economic in nature," or whether the causal chain is "too attenuated," remain open-textured, supplied by the LLM at the premise level.

In contrast to large language models, even when retrieval-augmented and chain-of-thought-prompted, the kernel is able to guarantee that no factor is missed or hallucinated and that there are no logical errors in applying the test.

C. Cross-Jurisdictional Sanction Proportionality

This third example illustrates the architecture's capacity for structural verification over open-textured, multi-regime legal analysis and the superintelligence thesis: that the system can traverse combinatorial legal spaces exceeding the capacity of any individual practitioner.

Consider the following scenario. A corporate client faces parallel sanctions exposure under EU Council Regulation 833/2014 and US Executive Order 13662. Defense counsel must identify, across the full corpus of ECtHR and CJEU jurisprudence, the closest factual analog in which a court held a restrictive measure disproportionate to its stated security objective, then map that ratio onto the client's fact pattern while simultaneously verifying that the proposed argument creates no additional exposure under OFAC's Specially Designated Nationals framework. In practice, a human team would partition this across sanctions, human-rights, and trade-compliance specialists, each conducting weeks of independent research before attempting synthesis.

[29] United States v. Lopez, 514 U.S. 549, 558–59 (1995); United States v. Morrison, 529 U.S. 598 (2000); Gonzales v. Raich, 545 U.S. 1 (2005).

Formally verified legal AI promises a different picture. At the autoformalization level, the LLM encodes the client's factual profile as a structured query over treaty bases, sanctioned-entity characteristics, and proportionality factors drawn from ECtHR doctrine. The kernel, then, can verify structural properties: that the cited precedent exists and its ratio is faithfully extracted; that the analogical mapping between the precedent's fact pattern and the client's is well-formed; that the proposed argument does not contradict the OFAC regulatory framework; and that the analytical sequence is complete – no stage of the proportionality test (legitimate aim, suitability, necessity, proportionality *stricto sensu*) is omitted. The explanation layer then produces a grounded memorandum: each analytical step is a triple pairing natural language with the specific ECtHR or CJEU citation and the structural mapping—an argument whose construction required simultaneous search across three legal regimes that no individual practitioner could have composed in comparable time.

The output is a structurally verified argument: complete, correctly sequenced, deductively valid in its scaffolding, with every citation grounded. The substantive persuasiveness of the proportionality claim remains a matter of practitioner judgment. But the structural guarantees already exceed what any existing legal AI system provides.

## IV — OBJECTIONS

One may object that the problem of open texture is so prevalent in law that any nontrivial legal question will escape the reach of formal verification. If so, the architecture is confined to trivial legal analyses, and the claim to legal superintelligence is empty.

This objection assumes that the boundary between bound determinations and open-textured analysis is fixed, with most nontrivial reasoning on the open-textured side. We challenge that assumption. The boundary is dynamic, and the mechanism that moves it is precedent.

Consider "economic in nature" in the *Lopez* test. In the abstract, the concept is open-textured. But after *Lopez*, *Morrison*, and *Raich*, a court asking whether the cultivation of marijuana for personal medical use is economic in nature is not exercising unconstrained discretion. It is bound by *stare decisis* to reach the conclusion that *Raich* reached. The open-textured concept has, for this class of facts, closed. This is the ordinary operation of the common law: in Hart's terminology, every time a court decides a penumbral case, it narrows the penumbra and expands the core. What was once open-textured becomes bound.[30]

The formally verifiable domain therefore includes not only purely computational provisions but also every open-textured concept that has been given determinate content by a sufficiently developed body of case law. And "sufficiently developed" is the norm in mature legal systems. Hart himself acknowledged that "the life of the law consists to a very large extent in the guidance both of officials and private individuals by determinate rules which, unlike the applications of variable standards, do not require from them a fresh judgment from case to case"[31]. The verifiable domain is therefore the vast territory in which most legal reasoning actually occurs.

[30] Our example mirrors the canonical illustration in the literature that is Horty's reconstruction of Stewart v. Dutra Constr. Co., 543 U.S. 481 (2005), and the precedent it turned on, DiGiovanni v. Traylor Bros., Inc., 959 F.2d 1119 (1st Cir. 1992). Before DiGiovanni classified a comparable construction platform as a non-vessel, the status of a dredge in *Stewart* was genuinely open; the court of appeals then treated the classification as settled by that precedent, with no recourse to the "core meaning" of "vessel." (The Supreme Court later reversed on the merits, holding the dredge a vessel.) See Horty, *The Logic of Precedent*, *supra* note 11, §1.3.2.

Of course, this assumes the relevant formalizations of the law and of the facts of the case to be sufficiently fine-grained to allow most of the facts encountered in practice to be determined; until this is the case, much nontrivial reasoning will remain on the open-textured side. But there is little reason to doubt that the proliferation of precedent and technology will at some point tip this scale.

A further objection targets the gap between the statute and its formalization. If the $\lambda_{law}$ program encodes the wrong rule, the kernel will faithfully verify a wrong conclusion. The guarantee is only as good as the formalization.

This is correct. The architecture does not eliminate the specification gap. It narrows it through three mechanisms. First, if the formalization disagrees with legal commentary on a worked example, the test fails. Second, tests can be automatically generated: concolic execution can generate hundreds of thousands of inputs exploring both branches of every default, catching edge cases that manual review would miss.[32] Third, for formalizations of statutory law, the system can use literate programming, where the $\lambda_{law}$ code sits side-by-side with the statutory text, enabling paragraph-level review by legal experts.

Furthermore, note that the issues that arise from the specification gap are strictly smaller than the challenges to existing AI systems. After all, they face the same specification problem: they must correctly identify and interpret the applicable rule and also face the problems of hallucination, miscalculation, and opaque reasoning.

## V — CONCLUSION

We argued that verification is the missing piece if AI is to achieve superhuman legal capabilities. To supply it, we adapted the *LLM proposes, verifier disposes* architecture that has reshaped mathematical AI: a typed legal calculus extending Catala, a small formally specified verification kernel combining type checking, static analysis, and SMT-backed assertion verification, and reinforcement learning from verifier feedback as the training signal for autoformalization.

For bound determinations, the architecture provides provable correctness: the system certifies, with deductive certainty, that the result follows from the formalized rules. For open-textured analysis, it provides structural verification: every required factor is addressed, in the correct order, with valid deductive links between steps. Reasonable lawyers may still disagree about the premises, and the architecture does not aspire to settle contested judgment. Neither guarantee is available from existing legal AI systems, and the value of both grows over time: as the formalized corpus expands, autoformalization improves, and recurrent patterns of application acquire determinate structure under sustained expert scrutiny.

What remains is the residue that verification cannot reach. The penumbra of open-textured standards will continue to require what it has always required: human judgment and choices that define the law of our societies. That is where legal reasoning is irreducibly human, and the architecture is designed to expose those moments rather than absorb them. Everything upstream—the selection of the governing test, the completeness of its application, the soundness of each deductive link, the grounding of each citation—is formally tractable, and it is there that the architecture will clear away preventable error. The path to a legal AI system worthy of the trust the profession demands runs through that division of labor, and no further.

[31] Hart, supra note 6, at 132.

[32] Pierre Goutagny, Aymeric Fromherz & Raphaël Monat, CUTECat: Concolic Execution for Computational Law, ESOP (2) 2025, at 31.